\documentclass[5p,times]{elsarticle}

\usepackage{amssymb}
\usepackage{amsmath,amsfonts}
\usepackage{algorithmic}
\usepackage{algorithm}
\usepackage{array}
\usepackage[caption=false,font=normalsize,labelfont=sf,textfont=sf]{subfig}
\usepackage{textcomp}
\usepackage{stfloats}
\usepackage{url}
\usepackage{verbatim}
\usepackage{graphicx}
\usepackage{mathrsfs}
\usepackage{multirow}
\usepackage{booktabs}

\journal{Nuclear Physics B}

\begin{document}

\begin{frontmatter}



\title{SMAP: A Novel Heterogeneous Information Framework for Scenario-based Optimal Model Assignment}


\author{Zekun Qiu}
\ead{qzk@buaa.edu.cn}
\author{Zhipu Xie \corref{cor1}}
\ead{xiezp12@chinaunicom.cn}
\author{Zehua Ji}
\ead{jizehua@buaa.edu.cn}
\author{Yuhao Mao}
\ead{maoyuhao@buaa.edu.cn}
\author{Ke Cheng}
\ead{ckpassenger@buaa.edu.cn}
\cortext[cor1]{Corresponding author}

\affiliation{organization={State Key Laboratory of Software Development Environment},
            addressline={Beihang University, Haidian District}, 
            city={Beijing},
            postcode={100191}, 
            country={China}}


\begin{abstract}
The increasing maturity of big data applications has led to a proliferation of models targeting the same objectives within the same scenarios and datasets.
However, selecting the most suitable model that considers model's features while taking specific requirements and constraints into account still poses a significant challenge.
Existing methods have focused on worker-task assignments based on crowdsourcing, they neglect the scenario-dataset-model assignment problem. 
To address this challenge, a new problem named the Scenario-based Optimal Model Assignment (SOMA) problem is introduced and a novel framework entitled Scenario and Model Associative percepts (SMAP) is developed.
SMAP is a heterogeneous information framework that can integrate various types of information to intelligently select a suitable dataset and allocate the optimal model for a specific scenario. 
To comprehensively evaluate models, a new score function that utilizes multi-head attention mechanisms is proposed. 
Moreover, a novel memory mechanism named the mnemonic center is developed to store the matched heterogeneous information and prevent duplicate matching.
Six popular traffic scenarios are selected as study cases and extensive experiments are conducted on a dataset to verify the effectiveness and efficiency of SMAP and the score function.

\end{abstract}




\end{frontmatter}






\section{Introduction}
The exponential growth of big data applications has prompted researchers to develop numerous application models to cater to diverse scenarios. For each scenario or dataset, a large number of  homogeneous models with comparable effects have been developed, including visual data detection models such as ResNet \cite{he2016deep} and MobileNet \cite{howard2017mobilenets} for ImageNet \cite{deng2009imagenet} and tabular data models like STDN \cite{yao2019revisiting} and TabNet \cite{arik2021tabnet} for NYCTaxi. Despite some of these models satisfying specific requirements, other essential factors and model characteristics are not adequately considered during the design process. For example, in deep neural networks, models with high accuracy can be inefficient due to their large number of layers, making them unsuitable for real-time applications. Additionally, some models' features such as limited citations and GitHub stars can hinder the recognition and acceptance of certain models. Therefore, it is crucial to select an appropriate model that balances factors such as accuracy, efficiency, and acceptability.  Identifying the best model under relevant criteria and constraints remains a critical research problem in computer science.

In the past few years, crowd-based matching \cite{yuen2011task,shu2018anonymous,wu2019bptm,10.1145/2588555.2588576} and machine learning-based matching methods \cite{doan2004ontology,rong2012machine,ristoski2018machine,berlin2002database} have been widely applied in recommendation systems and other fields.
Xing et al. \cite{xing2019multi} proposed a novel algorithm for multi-attribute crowdsourcing task assignment, which takes into consideration the various attributes of tasks as well as the interaction between tasks and workers. The algorithm aims to improve the efficiency and accuracy of task assignment in crowdsourcing platforms.
Zhao et al. \cite{zhao2022outlier} proposed an outlier detection method for streaming task assignment in crowdsourcing. The method employs a clustering-based outlier detection algorithm to identify and remove outlier workers in real-time, thereby enhancing the quality of task assignment.
Nottelmann et al. \cite{nottelmann2005information} presented a machine learning-based approach for database schema matching, which utilizes probabilistic methods to learn the similarities between different database schemas. The proposed method is effective in matching database schemas and can be applied to various applications that require schema matching.
Sahay et al. \cite{sahay2020schema} proposed a framework for schema matching based on machine learning techniques. The framework employs supervised learning to train classifiers that can determine the similarity between schema elements.
Although crowd-based and neural network-based matching methods have shown good performance in various fields, they overlook the relationships among the scenarios, datasets, and models. Therefore, they cannot be directly applied to optimal model matching based on the scenario.

\begin{figure*}[t]
	\centering
	\frenchspacing
\includegraphics[width=5.3in]{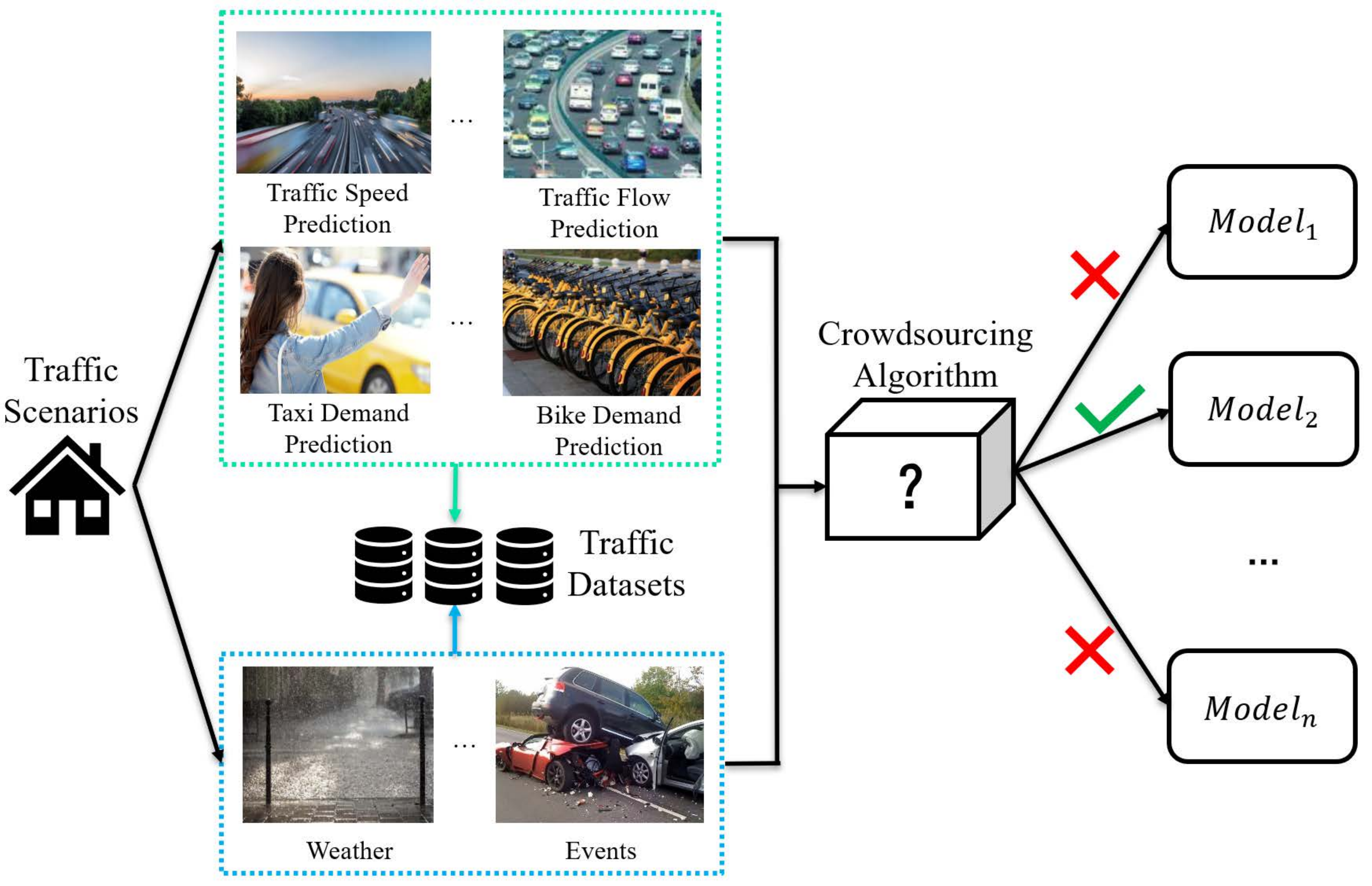}
\caption{An example of the optimal traffic model allocation in the crowdsourcing field.}
	\label{Introduction_1}
\end{figure*}

As shown in Fig.\ref{Introduction_1}, the traffic domain encompasses a multitude of scenarios, such as predicting traffic speed considering weather conditions, predicting event-driven traffic flow, and predicting taxi demand using point of interest (POI) information. The details are shown in the Appendix. For specific traffic scenarios, numerous traffic models have been developed, and extensive traffic datasets have been collected. For researchers in the transportation field, identifying the optimal model based on a suitable dataset for a particular traffic scenario may not be a challenging task. However, researchers from other disciplines may not possess the necessary professional knowledge to do so. Hence, designing an effective framework to assist researchers in easily identifying optimal models for any scenario on different datasets is essential.

Selecting the optimal model for datasets and scenarios presents significant challenges. These challenges are as follows:
(1) Evaluations of Models. The assessment of models by researchers is typically based on conventional metrics, yet there exists a tendency to neglect certain pragmatic factors. A holistic assessment of a model's performance must extend beyond the use of metrics alone; other critical factors like scenario-specific suitability and recognition by the public (such as citations and GitHub stars) must also be considered. Consequently, integrating the evaluation of model performance, adaptability to the scenario, and model features into a cohesive framework stands as a major challenge in model evaluation.
(2) Dataset Selection. The vast number of datasets collected makes selecting a suitable dataset for a particular scenario challenging. A suitable dataset can enhance the model's expressive capacity and shorten its training time. However, dataset selection is a difficult process due to the multitude of characteristics, including collection time, dataset type, number of downloads, geographic location, and time window.
(3) Memory. The allocation of an optimal model to a particular scenario forms a memory in the brain, enabling easy retrieval for subsequent encounters with the same scenario. The challenge is how to simulate memory in our problem.


To tackle these challenges, a new problem called the SOMA problem has been proposed, which aims to maximize the total scores to achieve the optimal model assignment. To solve this problem, a new framework called SMAP has been developed.
SMAP is a heterogeneous information framework that can integrate various types of information, such as scenarios, datasets, and models, and capture the relationships among them to intelligently select a suitable dataset and allocate the optimal model to a specific scenario. 
For the newly added scenarios in the scenario set, SMAP employs a selection process to identify datasets with desirable features and that are appropriate for given scenario types. Additionally, the framework selects candidate models from a set of models based on the scenario types to ensure their suitability for the specific scenario.
This process is akin to constructing a heterogeneous information network, where the matching of a dataset and a scenario can be likened to creating a link between the corresponding vertices in the heterogeneous information graph. Extensive experiments are implemented on the selected datasets to determine the performance of these models by utilizing several widely used metrics.
Furthermore, SMAP develops a score function to comprehensively evaluate these models by utilizing the multi-head attention mechanism. The output of the score function is the final score of each model.
An effective SOMA algorithm based on the greedy approach is presented to achieve the optimal model assignment for the suitable datasets and new scenarios, where the models' final scores can be regarded as the task utility in crowdsourcing.
A memory mechanism named the mnemonic center is developed to store the matched heterogeneous information and prevent duplicate matching.

In summary, the key contributions of this work are summarized as follows:

\begin{itemize}
\item An effective heterogeneous information framework named SMAP, which can integrate the heterogeneous information of scenarios, datasets and models as well as capturing the relationship among them, is proposed to intelligently select a suitable dataset based on its characteristics and allocate the optimal model to a special scenario by exploiting an effective SOMA algorithm. 
\item 
A score function is proposed to comprehensively evaluate models by utilizing the multi-head attention mechanism. To simulate brain activities in the real world, a new memory mechanism named the mnemonic center is developed to record the matched heterogeneous information to avoid duplicate matching.
\item  Six popular traffic scenarios are selected as study cases, and extensive experiments are conducted on a dataset to verify the effectiveness and efficiency of SMAP and the score function.
\end{itemize}
The remaining sections of this paper are structured as follows: In Section 2, a concise overview of the related work concerning both crowd-based and machine learning-based matching techniques is provided. Section 3 presents the problem definitions and formulations that are addressed by our approach. In Section 4, our proposed SMAP framework is described in detail, which includes the score function and SOMA algorithm. In Section 5, the experimental results of our approach for six popular traffic scenarios are introduced, and finally, in Section 6, a conclusion to our work is provided.
\section{Related Work}
\subsection{Crowd-based Matching}
Task matching is a critical aspect of crowdsourcing platforms that aims to assign the most suitable workers to tasks. Numerous approaches have been proposed to tackle this challenge, leveraging innovative techniques such as machine learning, blockchain, and privacy-preserving algorithms. Some of the research focuses on task matching and scheduling issues, intending to optimize the matching between tasks and workers, and improve work efficiency and quality.
Yuen et al. \cite{yuen2011task} proposed a framework for task assignment, aiming to establish efficient matches between task requesters and workers. The framework relies on a matching method that measures the compatibility between task features and worker skills. Deng et al. \cite{deng2015task} proposed a task matching and scheduling algorithm for multi-worker spatial crowdsourcing to improve task completion rates and worker profits. The algorithm takes into account worker locations and movement speeds, assigns tasks to the closest workers, and uses dynamic programming to determine the optimal task assignment and scheduling strategy. Xing et al. \cite{xing2019multi} proposed a new multi-attribute crowdsourcing task assignment algorithm that focuses on the multiple attribute characteristics of tasks and the interaction between tasks and workers. Yu et al. \cite{yu2019software} proposed a software crowdsourcing task allocation algorithm based on dynamic utility. The authors introduced a novel utility function that considers both the worker's performance and the task's urgency.

On the other hand, some research emphasizes privacy protection issues by using encryption techniques or technologies such as blockchain to protect the privacy of task publishers and workers, as well as ensuring the security of task data. Shu et al. \cite{shu2018anonymous} proposed a privacy-preserving task matching scheme based on anonymous authentication and secure multi-party computation to protect the privacy of task requesters and workers. Wu et al. \cite{wu2019bptm} proposed a privacy-preserving task matching scheme based on blockchain technology to protect the privacy of task requesters and workers. 
Kadadha et al. \cite{kadadha2021two} proposed a two-sided preferences task matching mechanism for  crowdsourcing platforms based on the blockchain. The authors propose a blockchain-based smart contract that allows both workers and requesters to express their preferences and constraints for task matching. 
Song et al. \cite{song2021privacy} proposed a privacy-preserving task matching scheme based on crowdsourcing. The scheme uses a threshold similarity search algorithm to match tasks and workers, and uses homomorphic encryption technology and reversible encryption technology to protect the privacy of tasks and workers. 
Wang et al. \cite{wang2021reliable} propose a reliable and privacy-preserving task matching mechanism for crowdsourcing on the basis of the blockchain. The authors introduce a blockchain-based task matching protocol that guarantees the reliability and privacy of task matching. Guo et al. \cite{guo2020fedcrowd} propose a federated and privacy-preserving crowdsourcing platform based on blockchain technology.

In addition, some research primarily focuses on data matching and quality issues, aiming to improve the efficiency and accuracy of data matching and cleaning, as well as enhancing data quality.
Gokhale et al. \cite{10.1145/2588555.2588576} proposed a novel method that harnesses machine learning and semi-supervised learning techniques to automate entity matching tasks in crowdsourcing platforms. The proposed approach obviates the need for task requesters to manually identify and assign tasks to workers, and instead employs a sub-task division and assignment strategy. Zhang et al. \cite{zhang2013reducing} developed a crowdsourcing-based method to mitigate the uncertainty in schema matching. This method leverages the expertise of workers in annotating data tuples to facilitate the identification of the corresponding matching relationship. Additionally, Zhao \cite{zhao2022outlier} introduced an outlier detection algorithm based on clustering that identifies and eliminates anomalous workers in real-time, thereby improving the accuracy and efficiency of task assignments in streaming crowdsourcing scenarios. In conclusion, these studies underscore the efficacy of crowdsourcing in improving the quality and productivity of entity matching tasks. Nonetheless, their focus primarily concerns task-worker aspects, which may limit their utility in addressing scenario-dataset-model challenges such as those encountered in SMAP.

\subsection{Machine Learning-based Matching}
The application of machine learning has gained widespread recognition in various matching tasks, including but not limited to ontology matching, instance matching, product matching, database schema matching, and address matching. This section provides an overview of several related works in these domains.
Doan et al. \cite{doan2004ontology} proposed a machine learning-based ontology matching method, which utilizes the random forest algorithm and employs various feature selection techniques such as information gain, correlation, and chi-squared tests.
Jaber et al. \cite{jaber2022machine} proposed a machine learning-based semantic pattern matching model for the registration of remote sensing data. Similarly, 
Comber et al. \cite{comber2019machine} compared the performance of two machine learning methods, namely word2vec and conditional random field, for address matching. Xiang et al. \cite{xiang2022design} presented an intelligent education system resource matching model that was based on machine learning and achieves the matching and allocation of educational resources by learning the mutual relationships between them. 
Moreover, Nottelmann et al. \cite{nottelmann2005information} proposed a machine learning and probabilistic methods-based database schema matching method that learned similarities between schema elements to match database schemas. 
Sahay et al. \cite{sahay2020schema} proposed a framework for schema matching that utilized supervised learning to train classifiers capable of determining the similarity between schema elements. 
Lastly, Paganelli et al. \cite{paganelli2021automated} proposed an automated approach to entity matching utilizing machine learning techniques, which involved feature engineering and model selection.

In similarity-based matching, Rong et al. \cite{rong2012machine} introduced a machine learning approach based on similarity metrics, which incorporates several commonly used similarity metrics such as cosine similarity, Euclidean distance, and Manhattan distance, as well as feature selection and dimensionality reduction techniques.
Ristoski et al. \cite{ristoski2018machine} proposed a machine learning-based approach for product matching and classification, which uses multiple feature selection and dimensionality reduction techniques, such as principal component analysis and linear discriminant analysis.
Berlin et al. \cite{berlin2002database} proposed a machine learning-based method for database schema matching, which involves multiple feature selection and dimensionality reduction techniques, including principal component analysis (PCA) and linear discriminant analysis (LDA).
Yu et al. \cite{yu2015machine} proposed a method for physical design using machine learning and pattern matching techniques, which includes several common classification algorithms and dimensionality reduction techniques.

The primary objective of our framework is to provide a resolution for optimal model matching based on contextual scenarios. The conventional machine learning techniques may not be adequate to tackle this issue. Moreover, several of the above-discussed approaches are tailored to specific use cases, making their generalization to our problem domain difficult.

\begin{figure}[t]
	\centering
	\frenchspacing
\includegraphics[width=3.4in]{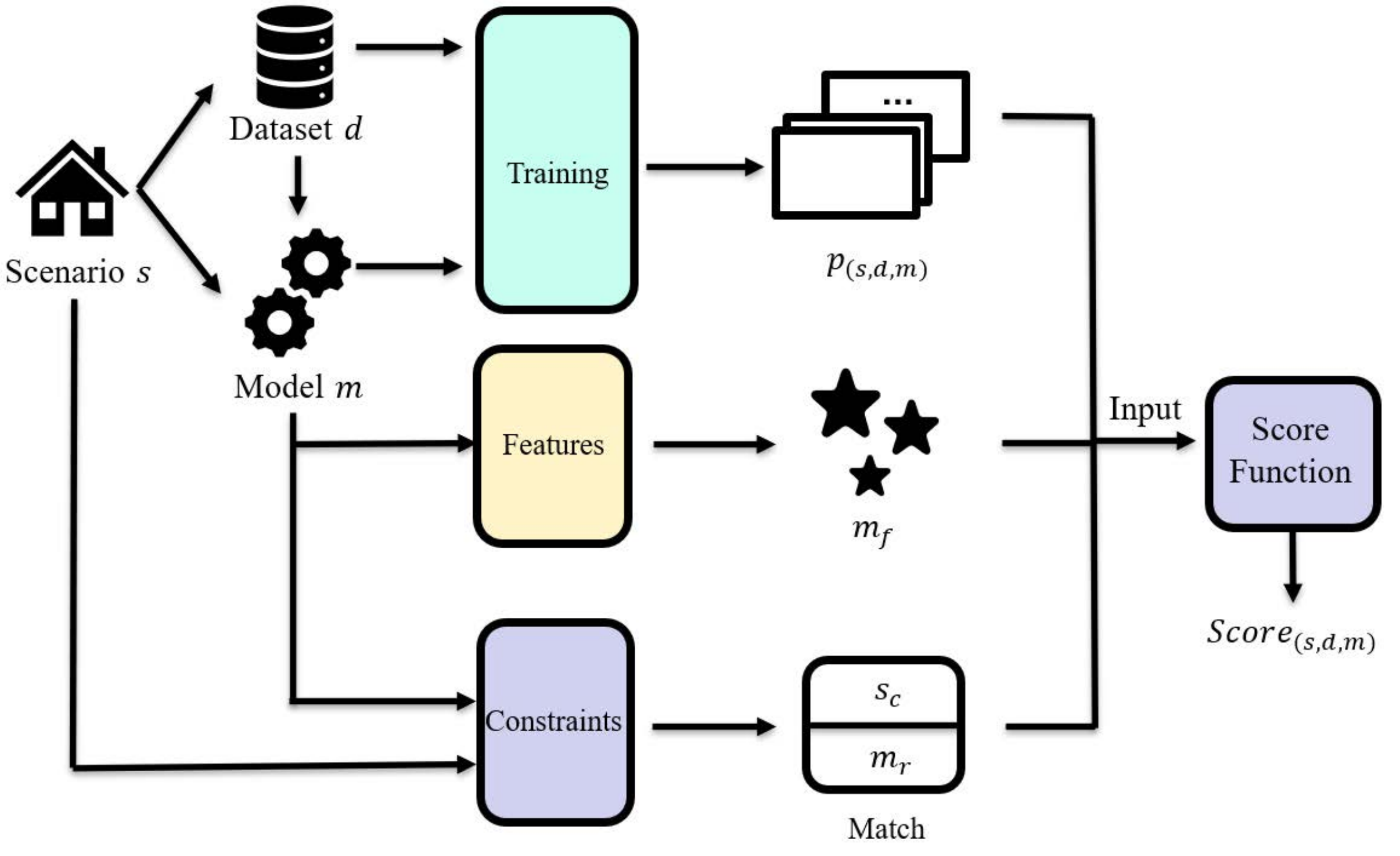}
\caption{The core idea of the score function.}
	\label{scorefunction}
\end{figure}

\section{Problem Definitions}
In this section, the SOMA problem is defined and several constraints are described. The details are shown as follows:

\textbf{Definition 1} (Scenario). A scenario is described as a contextual framework that characterizes the environment and constraints under which a machine learning task is executed. A scenario $s$ can be represented as a tuple $s=(s_t,s_c)$, where $s_t$ represents the scenario type and $s_c=(s_{c_1},s_{c_2},\ldots,s_{c_n})$ represents the scenario constraints. The set of all scenarios is denoted as $\mathcal{S}$.



\textbf{Definition 2} (Dataset). A dataset is a collection of data within a specific scenario, which can be represented as a tuple $d=(d_t,d_f)$, where $d_t$ represents the dataset type and $d_f=(d_{f_1},d_{f_2},\ldots,d_{f_n})$ represents the dataset features. The set of all datasets is denoted as $\mathcal{D}$.


\textbf{Definition 3} (Model). 
A model is usually constructed by researchers in a particular field using a dataset as their training or inference sample, with the purpose of solving a specific classification or regression problem. A model $m$ can be represented as a tuple $m=(m_t,m_f,m_r)$, where $m_t$, $m_f = (m_{f_{1}},m_{f_{2}},...,m_{f_{n}})$ and $m_r = (m_{r_{1}},m_{r_{2}},...,m_{r_{n}})$ are the type, features and requirements of the model, respectively. The set of all models is denoted as $\mathcal{M}$.

\textbf{Definition 4} (Performance). The performance of a model refers to the level of accuracy it can achieve on the test set. When discussing a model’s performance, it is essential to specify the corresponding scenario and dataset. The performance of model $m$ on dataset $d$ for scenario $s$ is typically evaluated using multiple predefined evaluation metrics. Therefore, ${p}_{(s,d,m)}$ can be represented as a tuple, i.e., ${p}_{(s,d,m)} = ({p_1}_{(s,d,m)}, {p_2}_{(s,d,m)},…,$\par \noindent$ {p_k}_{(s,d,m)})$, where ${p_i}_{(s,d,m)}$ represents the $i^{th}$ metric.

\textbf{Definition 5} (Score Function). The score of a model on a suitable dataset for a special scenario is measured by a score function $\mathcal F$:
\begin{equation}\label{formula}
\begin{split}
score_{(s,d,m)}= \mathcal {F}(\mathbb{I} (match(s_c,m_r)),p_{(s,d,m)}, m_f).
\end{split}
\end{equation}

Where, $\mathbb{I}(match(s_c,m_r))$ returns $1$ if the scenario’s context $s_c$ matches the model’s requirements $m_r$, and $0$ otherwise. This term acts as a binary indicator variable, indicating whether the model is suitable for the given scenario or not. $p_{(s,d,m)}$ represents the performance of model $m$ on dataset $d$ for scenario $s$. And $m_f$ denotes model features that could impact its performance.
The core idea of the score function are shown in Fig. \ref{scorefunction}.




\textbf{Definition 6} (SOMA Problem). Given a set of scenarios $S \in \mathcal{S}$, a set of dataset $D \in \mathcal{D}$, a set of models $M \in \mathcal{M}$, and a score function $\mathcal F$, the SOMA problem aims to find an optimal allocation $A$ between the scenarios and models to maximize the total scores $MaxScore(A)=\sum_{s \in S, m \in M} \mathop{max}\limits_{d \in D}(score_{(s,d,m)})$. However, the optimal allocation must satisfy the following constraints:

\begin{figure*}[t]
	\centering
	\frenchspacing
\includegraphics[width=5.3in]{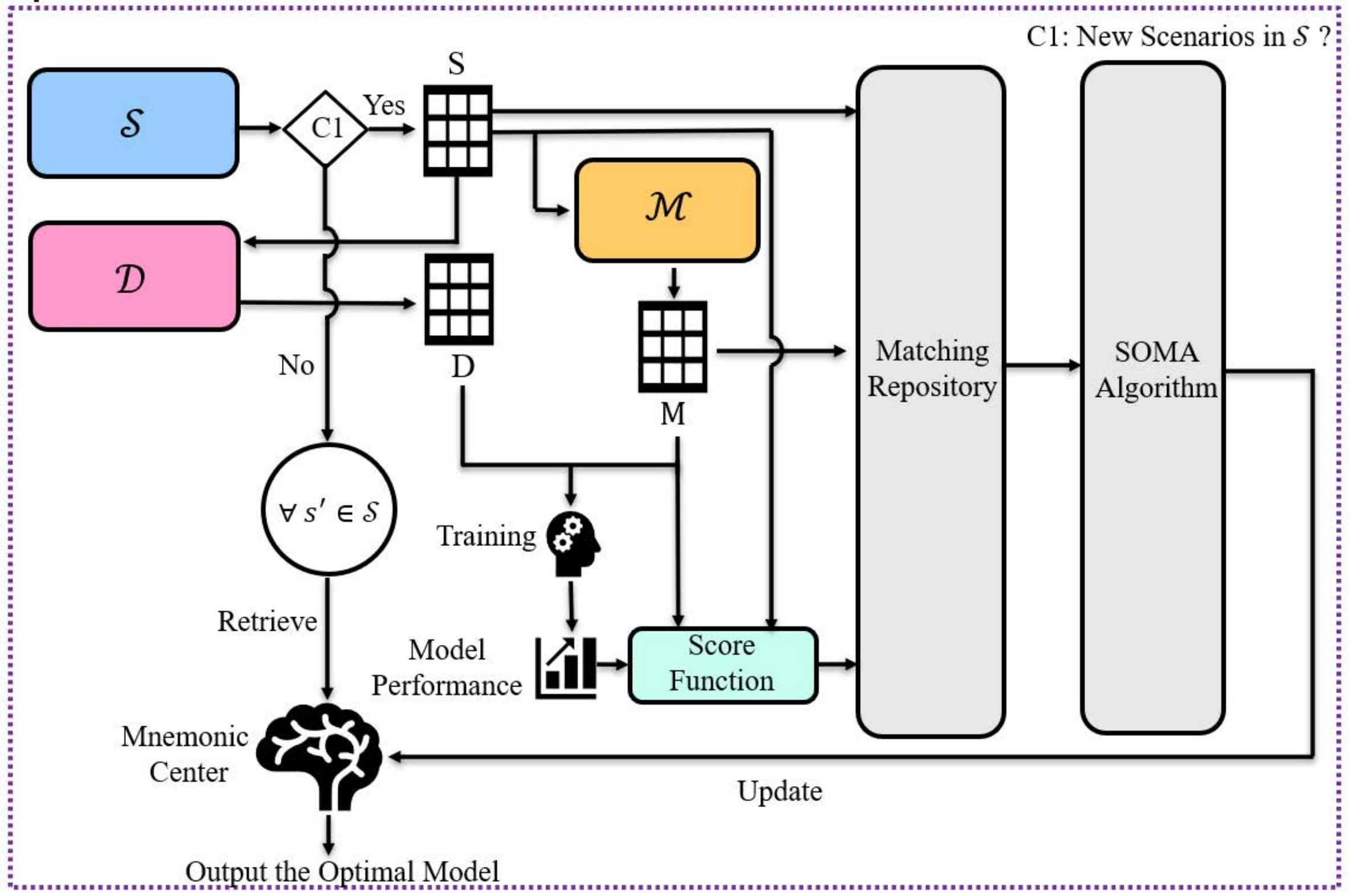}
\caption{The heterogeneous information framework SMAP.}
	\label{SMAP}
\end{figure*}

\begin{itemize}
\item {\verb|Invariable Constraint|}: Once a model $m$ is allocated to a scenario $s$, the allocation cannot be changed.
\item {\verb|Capacity Constraint|}: A scenario can only accept an optimal model, but an optimal model can be assigned to multiple  scenarios.
\item {\verb|Dataset Constraint|}: For a special scenario $s \in S$, there exists a $d \in D$ such that $d_t$ matches $s_t$.
\item {\verb|Model Constraint|}: For a special scenario $s \in S$, there exists an $m \in M$ such that $m_t$ matches $s_t$ at least.
\end{itemize}

\section{SMAP} \label{section_SMAP}
To solve the SOMA problem, a novel heterogeneous information framework, entitled SMAP, is developed, which is shown in Fig. \ref{SMAP}.  SMAP includes three types of information, i.e., scenarios, datasets and models. Instead of constructing the heterogeneous graph and learning the representation for heterogeneous nodes, rich characteristics of them is simply utilized, an effective score function is designed, and an algorithm is exploited to tackle the SOMA problem. Specially, SMAP is composed of an $\mathcal{S}$, a $\mathcal{D}$, an $\mathcal{M}$, a matching repository, a score function, an SOMA algorithm, and a mnemonic center. All scenarios, datasets and models are saved in the $\mathcal{S}$, $\mathcal{D}$ and $\mathcal{M}$, respectively. If the constraints of scenarios do not match the requirements of models, the score function will be $0$. Otherwise, the score function comprehensively evaluates models according to their features (i.e., the number of citations of each published paper and the number of stars on Github) and performance. The SOMA algorithm is a bipartite matching algorithm that can achieve the optimal model allocation.
SMAP first retrieves the $\mathcal{S}$ to judge whether there are new scenarios. If no new scenario is added, for any scenario in the $\mathcal{S}$, SMAP retrieves the mnemonic center and directly output the optimal model assigned previously. If new scenarios $S$ are presented in the $\mathcal{S}$, SMAP selects the suitable datasets $D$ from the $\mathcal{D}$ based on their features and scenario types and chooses candidate models $M$ from the $\mathcal{M}$ based on the scenario types. The selection process can refer to the dataset and model constraints in Definition 6.
SMAP regards datasets that satisfy the corresponding constraint with high number of downloads as suitable datasets.
Then, $p_{(s,d,m)}$ is obtained as the performance of model $m$ on dataset $d$ for scenario $s$ through extensive experiments by utilizing widely used metrics. 
The output of the score function $score_{(s,d,m)}$ represents the final score of model $m$ on dataset $d$ for scenario $s$, which can be regarded as the task utility in crowdsourcing.  
The inputs of the SOMA algorithm are $S$, $D$, $M$, and $score$, and the output of the SOMA algorithm is the optimal allocation $A$. In addition, the assigned scenario-dataset and model pairs are updated for the mnemonic center. When the same scenario is encountered, SMAP can directly retrieve the mnemonic center without rematching.
 
\begin{figure*}[t]
	\centering
	\frenchspacing
\includegraphics[width=5.3in]{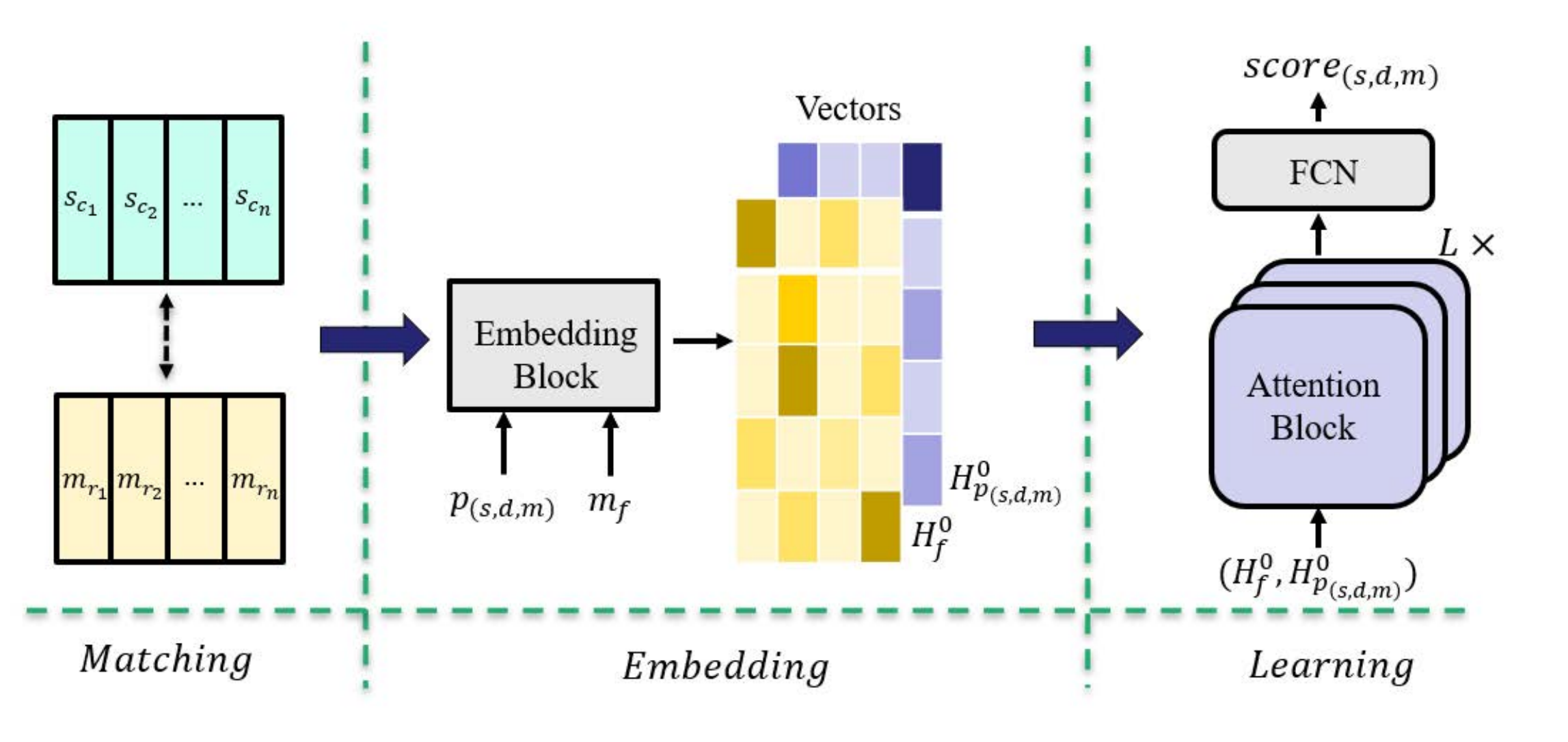}
\caption{Overall architecture of the score function.}
	\label{Attention1}
\end{figure*}

\subsection{Score Function}\label{score_function}
Attention mechanisms \cite{article} are efficient methods that can accurately filter out high-value features from a large amount of information. Attention mechanisms solve the difficulty of obtaining a reasonable vector representation when the input sequence of RNNs and LSTMs is long. Due to their effectiveness, attention mechanisms have been widely applied to various domains. However, they also have some disadvantages. When encoding information, attention mechanisms attend too much to their own positions. To overcome this limitation, a multi-head attention mechanism \cite{NIPS2017_3f5ee243} is developed to allow the model to jointly focus on more encoding representation information from different subspaces. 
In this paper, given the $s_{c}$, $m_{r}$, $m_{f}$, and $p_{(s,d,m)}$ of model $m$ on dataset $d$ for scenario $s$, the score function utilizes the multi-head attention mechanism, which focus on the most relevant features to extract hidden correlations and output the final score $score_{(s,d,m)}$. 


As shown in Fig. \ref{Attention1}, the overall framework of the score function comprises a matching function, an embedding block, a fully connected neural network (FCN) and a stack of $L$ attention blocks. A FCN is a special artificial neural network, where any node in the $(n-1)^{th}$ layer is connected to all nodes in the $n^{th}$ layer.
In SMAP, the matching function is first utilized to verify whether the candidate model is suitable for the scenario within the constraints. If $m_{r}$ does not match $s_{c}$, the matching function returns 0 and the score $score_{(s,d,m)}$ of model $m$ on dataset $d$ for scenario $s$ also equals 0.
Then, the embedding block is utilized to convert the input features into low-dimensional vectors. The output of the embedding block is represented as $H^{0}_{(s,d,m)}=\{H_{f}^0, H_{p_{(s,d,m)}}^0\}$. When the last attention block outputs the hidden states, an FCN is exploited to aggregate these correlations to obtain the final scores. All attention blocks produce outputs of $D$ dimensions. The formulas are shown as follows: 

\begin{equation}\label{FC1}
\begin{split}
score_{(s,d,m)} = ReLU(ReLU(H^{L}_{(s,d,m)}W_{1}+b_{1})W_{2}+b_{2}).
\end{split}
\end{equation}

$H^{L}_{(s,d,m)} \in R^{N \times D}$ is the output of the last attention block in model $m$ on dataset $d$ for scenario $s$, where $N$ is the size of input features of the first attention block.  $W_{1}, W_{2}, b_{1}$, and $b_{2}$ are learnable parameters. ReLU \cite{10.5555/3104322.3104425} is the activation function.

\begin{figure*}[t]
	\centering
	\frenchspacing
\includegraphics[width=5.2in]{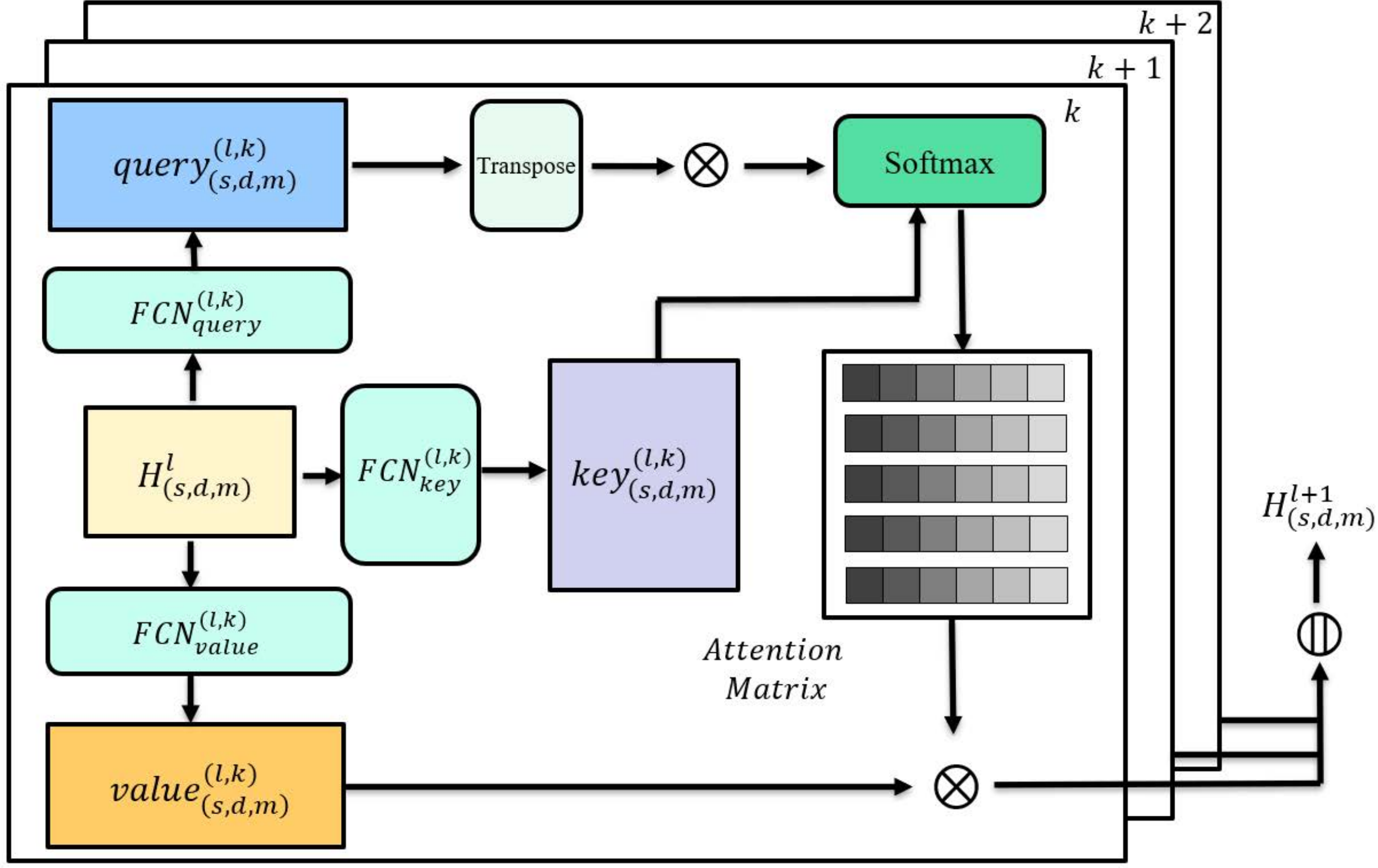}
\caption{The architecture of the attention block.}
	\label{Attention2}
\end{figure*}

To extract the hidden dependencies among different features, the attention blocks, each of which exploits the multi-head attention mechanism, are developed.
The input of the $l^{th}$ block is denoted as $H^{l}$, where the hidden state of model $m$ on dataset $d$ for scenario $s$ is represented as $H^{l}_{(s,d,m)}$.
Three full connection networks, each of which has only one hidden layer, are first exploited to construct the query vectors $\mathbf{query}^{(l,k)}_{(s,d,m)}$, key vectors $\mathbf{key}^{(l,k)}_{(s,d,m)}$ and value vectors $\mathbf{value}^{(l,k)}_{(s,d,m)}$ of the $k^{th}$ head in the $l^{th}$ block. The dimensions of these query, key, and value vectors are denoted as $d_{query}$, $d_{key}$, and $d_{value}$, respectively, where $d_{query} = d_{key} = d_{value} = D / K$ and $K$ is the number of attention heads. The formulas are shown as follows: 

\begin{equation}\label{QKV}
\begin{split}
\mathbf{query}^{(l,k)}_{(s,d,m)}= FCN^{(l,k)}_{query}(H^{l}_{(s,d,m)}), \\
\mathbf{key}^{(l,k)}_{(s,d,m)}= FCN^{(l,k)}_{key}(H^{l}_{(s,d,m)}), \\
\mathbf{value}^{(l,k)}_{(s,d,m)}= FCN^{(l,k)}_{value}(H^{l}_{(s,d,m)}), \\
\end{split}
\end{equation}

Where, $FCN^{(l,k)}_{query}$, $FCN^{(l,k)}_{key}$, and $FCN^{(l,k)}_{value}$ represent three different full connection networks in the $k^{th}$ head, each of which has only one hidden layer.

The dot-product approach \cite{NIPS2017_3f5ee243} is utilized to discover the differences and capture the relevance between the query and key vectors. For convenient expression, the $i^{th}$ and $j^{th}$ features are taken as examples to illustrate the whole calculation process. Their formulas are shown as follows: 

\begin{equation}\label{s1}
\begin{split}
&S_{i,j}^{k}=\frac{\left\langle {(\mathbf{query}_i)}_{(s,d,m)}^{l,(k)}, {(\mathbf{key}_j)}_{(s,d,m)}^{l,(k)}\right\rangle}{\sqrt{d_{query}}},
\end{split}
\end{equation}
where $\langle\cdot, \cdot\rangle$ represents the inner product operator and $d_{query}$ is the dimension of ${(\mathbf{query}_p)}_{s,d,m}^{l,(k)}$. The attention score in the $k^{th}$ head between the $i^{th}$ and $j^{th}$ features is computed as:

\begin{equation}\label{atten1}
\begin{split}
\alpha_{i, j}^{k}=\frac{\exp \left(S_{i, j}^{k}\right)}{\sum_{f \in N_{f}} \exp \left(S_{i, f}^{k}\right)}.
\end{split}
\end{equation}
where $N_f$ is a set containing all input features of the first attention block. The hidden states of the $i^{th}$ feature in the $(l+1)^{th}$ block are computed as follows:

\begin{equation}\label{atten2}
\begin{split}
{(H_i)}_{(s,d,m)}^{l+1}=\|_{k=1}^{K}\sum_{f\in N_f} \alpha_{i, f}^{k} \cdot {(\mathbf{value}_f)}_{(s,d,m)}^{l,(k)}.
\end{split}
\end{equation}

Where $\mathbf{value}$ is the value vector. $\|$ is the concatenation operation.

\begin{algorithm}[tb]
\caption{SOMA}
\label{Algorithm1}
\textbf{Goal}: $MaxScore(A)=\sum_{s \in  S, m \in  M} \mathop{max}\limits_{d \in  D}(score_{(s,d,m)})$\\
\textbf{Input}: Scenarios $S$, Dataset $D$, models $M$, $score$\\
\textbf{Output}: A feasible allocation $A$
\begin{algorithmic}[1] 
\STATE $A \gets \emptyset $ ;
\WHILE{True}
\STATE $((s,d), m) \gets$ A feasible pair of a scenario and model, along with a selected suitable dataset, that satisfies the constraints with the highest $score_{(s,d,m)}$; 
\IF {$((s,d), m) \; exists$}
\STATE $A \gets A \cup \{((s,d), m)\}$;
\ELSE
\STATE break;
\ENDIF
\ENDWHILE
\RETURN $A$;
\end{algorithmic}
\end{algorithm}






\subsection{SOMA Algorithm}
After obtaining the final scores of models on suitable datasets for special scenarios, the SOMA algorithm is exploited based on the greedy approach to achieve the optimal model assignment. 
The greedy approach is a generic method that divides the solution process into several steps to find the optimal solution of a problem. Each step employs the greedy principle to achieve local optimization so that the final stacked result is also optimal.
Algorithm \ref{Algorithm1} illustrates the procedure of SOMA. The inputs of the SOMA algorithm are the newly added traffic scenarios $S$, the select suitable dataset $D$, the candidate models $M$, and the model scores $score$ from the matching repository. The output is the optimal allocation $A$. In lines 2-9 , the unmatched edge that satisfies the constraints with the highest $score_{(s,d,m)}$ is iteratively added to $A$, if such an edge exists. If no such edge exists, the algorithm breaks the loop and returns the final allocation.

\begin{table}[t]
\begin{center}
\newcommand{\tabincell}[2]{\begin{tabular}{@{}#1@{}}#2\end{tabular}}
\caption{Examples of traffic datasets in the dataset set}\label{TD}
\scalebox{0.68}{
\begin{tabular}{ccccc}
\hline
Names&Types&Collection times&Locations&External factors\\
\hline
PEMSD8& \tabincell{c}{Speed, flow, \\occupancy}& \tabincell{c}{2016.7.1 -\\ 2016.8.31}& \tabincell{c}{San Bernardino \\ Area, USA}&-\\
\hline
TaxiBJ& Taxi GPS& 2013 - 2016& \tabincell{c}{Beijing,\\China}& \tabincell{c}{Weather, \\holidays}\\
\hline
TaxiNYC& Taxi trips& \tabincell{c}{2009 - present}& \tabincell{c}{New York, USA}& Holidays\\
\hline
PEMSD3& Traffic flow& \tabincell{c}{2018.9.1 -\\ 2018.11.30}& \tabincell{c}{District 3 of \\California, USA}&-\\
\hline
\tabincell{c}{Subway \\Transaction \\Dataset}& \tabincell{c}{Transaction \\ records}& \tabincell{c}{2016.6.1 - \\2016.6.29}& \tabincell{c}{Beijing,\\China}&-\\
\hline
\tabincell{c}{Bus Transaction \\Dataset}& Bus records& \tabincell{c}{2016.6.1 -\\ 2016.6.29}& \tabincell{c}{Beijing,\\China}&-\\
\hline
Didi\_HK2017& \tabincell{c}{Ride-hailing \\orders}&\tabincell{c}{2017.5.1 -\\ 2017.10.31}& \tabincell{c}{Haikou, \\China}&- \\
\hline
Didi\_BJ2017& \tabincell{c}{Ride-hailing \\orders}&\tabincell{c}{2017.3.1 -\\ 2017.12.31}& \tabincell{c}{Beijing, \\China}& POIs\\
\hline
\end{tabular}}
\end{center}
\end{table}

\section{Experiment}
In this section, six popular traffic scenarios are selected as study cases, i.e., traffic speed prediction, road traffic flow prediction, station-level subway passenger flow prediction, station-level bus passenger flow prediction, taxi demand prediction and ride-hailing demand prediction. 
Each scenario's constraints are set to prioritize performance. The citations of the published papers of models and the number of Github stars are selected as model features.
Moreover, these traffic scenarios are added to the $\mathcal{S}$ as new scenarios to evaluate the effectiveness of our proposed SMAP.

\subsection{Dataset}
The traffic datasets in $\mathcal{D}$ that used to evaluate the performance of traffic models are of high-quality, and most of them are publicly available. These datasets are collect from LibCity \cite{10.1145/3474717.3483923}, PaddlePaddle \cite{ma2019paddlepaddle} and Github, which cover millions of kilometers and wide time ranges. Table \ref{TD} shows some traffic datasets in $\mathcal{D}$, the attributes of which include their names, types, collection times, locations and external factors. The score function and SOMA algorithm are evaluated using a dataset that includes records containing information on the characteristics, requirements and performance of various traffic models. Each record comprises the model name, published paper, requirements, performance, the name of the experimental dataset, prediction process time length, number of citations and Github stars, external factors considered, traffic scenarios to which the model belongs, and a score that represents the model's ranking on the dataset for the corresponding traffic scenario. To establish the ground truth for the score, a team of workers are organized to label each record based on the characteristics and performance of the model. The MAE, MAPE, and RMSE, three widely used metrics, are exploited to evaluate the performance of  traffic models.

\subsection{Data Preprocessing}
After selecting the suitable traffic datasets and candidate models from $\mathcal{D}$ and the $\mathcal{M}$, respectively, based on the newly added traffic scenarios in $\mathcal{S}$, extensive experiments are conducted to determine the model performance. The same data preprocessing procedures are adopted as those in the published papers of the traffic models. In the score function experiments, Z-score normalization is exploited to standardize the data inputs. 50\%, 10\%, and 40\% of the data are utilized for training, validation and testing, respectively.

\subsection{Experimental Setting}
All experiments are conducted on a Linux server (CPU: Intel(R) Xeon(R) CPU E5-2667 v4 @ 3.20GHz; GPU: NVIDIA Titan Xp). 
For the experiments that evaluate the performance of the traffic models selected from the model set, the hyper-parameter settings and optimizer are the same as those in the published papers. 
For the experiments involving the score function, three hyper-parameters influence the model complexity, i.e., the number of attention heads $K$, the number of attention blocks $L$, and the dimensionality $d$ of each attention head. These hyper-parameters are tuned on the validation dataset, and the best performance is observed under the settings of $K=8$, $L=1$, and $d=8$. In addition, the score function adopting Adam optimizer \cite{Adam2015} is trained with an initial learning rate of 0.025. The batch size $BS$ is set to $64$.

\subsection{Evaluation Metrics}
Three widely adopted evaluation metrics, namely Hit@1, Hit@3, and Hit@5, are exploited to assess the effectiveness of SMAP. These metrics represent the average probability of hitting the ground truth from the top-1, -3, and -5 models allocated by SMAP, respectively, for each of the six traffic study cases. Hit@1 can be obtained using the SOMA algorithm as described earlier. To obtain Hit@3 and Hit@5, the SOMA algorithm is fine-tuned to recommend the top-3 and top-5 models, respectively, for each study case.

\begin{table*}[t]
\begin{center}
\caption{Performance comparison among the score function and baseline models.}\label{exp_result}
\scalebox{0.9}{
\begin{tabular}{c|ccc|ccc|ccc}
        \hline
		\multirow{2}{*}{Score Function} & 
		\multicolumn{3}{c|}{Exp 1-10}  &
		\multicolumn{3}{c|}{Exp 11-20}  &
		\multicolumn{3}{|c}{Exp 21-30}  \\
		\cline{2-10}
		 & Hit@1& Hit@3&Hit@5& Hit@1& Hit@3&Hit@5& Hit@1& Hit@3&Hit@5 \\
		\hline
        SVD&0.38&0.58&0.58&0.38&0.65&0.65&0.38&0.62&0.62\\
        SVD++&0.38&0.63&0.78&0.40&0.73&0.83&0.42&0.70&0.83\\
        NMF&0.37&0.62&0.72&0.42&0.67&0.75&0.42&0.63&0.78\\
        Slope One&0.47&0.50&0.50&0.46&0.48&0.48&0.43&0.50&0.50\\
        Co-Clustering&0.42&0.65&0.73&0.43&0.62&0.73&0.37&0.63&0.83\\
        AdaBoost&0.65&\textbf{0.98}&\textbf{1.00}&0.68&\textbf{0.93}&\textbf{0.98}&0.68&0.92&\textbf{0.95}\\
        Attention&\textbf{0.83}&0.83&0.98&\textbf{0.82}&0.87&0.88&\textbf{0.85}&\textbf{0.93}&0.93\\
        \hline
        \multirow{2}{*}{Score Function} &
        \multicolumn{3}{c|}{Exp 31-40}  &
        \multicolumn{3}{c}{Exp 41-50} &
        \multicolumn{3}{|c}{Exp 1-50} \\
        \cline{2-10}
        &Hit@1& Hit@3&Hit@5& Hit@1& Hit@3&Hit@5&Hit@1&Hit@3&Hit@5 \\
        \hline
        SVD&0.43&0.65&0.65&0.43&0.68&0.68&0.40&0.64&0.64\\
        SVD++&0.45&0.72&0.80&0.43&0.67&0.80&0.42&0.69&0.81\\
        NMF&0.42&0.73&0.77&0.45&0.70&0.83&0.41&0.67&0.77\\
        Slope One&0.50&0.5&0.5&0.45&0.47&0.47&0.46&0.49&0.49\\
        Co-Clustering&0.37&0.63&0.75&0.40&0.62&0.77&0.40&0.64&0.76\\
        AdaBoost&0.70&\textbf{0.93}&\textbf{0.97}&0.67&0.88&0.97&0.68&\textbf{0.9}3&\textbf{0.97}\\
        Attention&\textbf{0.75}&0.85&0.95&\textbf{0.82}&\textbf{0.92}&\textbf{0.97}&\textbf{0.81}&0.91&0.94\\
        \hline
\end{tabular}}
\end{center}
\end{table*}

\subsection{Experimental Results}
\subsubsection{Forecasting Performance Comparison}
The score function is evaluated in comparison with several classical methods. The results demonstrate that the score function has superior performance. The baselines are shown as follows:
\begin{itemize}
\item[$\bullet$] \textbf{SVD} \cite{SVD}: Singular Value Decomposition (SVD) is a matrix factorization algorithm that can decompose a matrix into the product of multiple matrices. In the field of recommendation systems, the SVD algorithm is widely used in collaborative filtering recommendations.
\item[$\bullet$] \textbf{SVD++} \cite{SVD++}: SVD++ is a collaborative filtering recommendation algorithm that extends the SVD algorithm and is used to predict user ratings for items in a recommendation system. The algorithm utilizes both an implicit feedback matrix and an explicit rating matrix to obtain latent feature vectors for users and items through optimization algorithms.
\item[$\bullet$] \textbf{NMF} \cite{NMF}: Non-negative matrix factorization (NMF) is a matrix factorization algorithm that aims to map high-dimensional data to a lower-dimensional space through the use of a non-negative matrix to describe the data’s structure and features. Unlike the SVD algorithm, NMF is a non-Euclidean distance measurement method.
\item[$\bullet$] \textbf{Slope One} \cite{lemire2008slope}: The Slope One algorithm is an efficient recommendation algorithm that builds a weighted rating deviation model based on a user’s historical rating data. This model is then used to predict a user’s rating for unrated items. The Slope One algorithm is often applied to scenarios with sparse or large datasets, demonstrating excellent performance and scalability.
\item[$\bullet$] \textbf{Co-Clustering} \cite{NIPS2007_d56b9fc4}: The Co-Clustering algorithm is a cluster-based recommendation algorithm that performs simultaneous clustering of users and items. It assigns items within and between clusters to generate recommendations. The Co-Clustering algorithm is capable of handling high-dimensional and sparse data, as well as data containing multiple dimensions while preserving their respective attributes.
\item[$\bullet$] \textbf{AdaBoost} \cite{FREUND1997119}: AdaBoost, short for adaptive boosting, is an ensemble learning method that was initially developed to improve the performance of binary classifiers. Similar to decision trees, AdaBoost can also be used for regression tasks; this approach adopts an iterative method to turn weak classifiers into strong ones by capturing mistakes and learning from them.
\end{itemize}


Every experiment is repeated 50 times and report the average performance in Table \ref{exp_result}. The results are partitioned into six groups according to their experimental frequency. The first five groups represent the average results obtained every 10 experiments, and the last group contains the average results of 50 experiments. 
Although SMAP aims to allocate the optimal model for a new scenario, it is impossible for any algorithm to guarantee a 100\% Hit@1. To more effectively evaluate the score function in SMAP, the SOMA algorithm is fine-tuned to calculate the average probability of accurately hitting the ground truth among the top 3 and top 5 models allocated by SMAP for each study case. These metrics are represented by Hit@3 and Hit@5, respectively.
In the most critical metric of Hit@1, our model shows superior performance compared to other baseline methods. This means that when SMAP performs model allocation tasks in new scenarios, the score function represented by the multi-head attention mechanism has the highest probability of assigning the optimal model to these scenarios. In the Hit@3 and Hit@5 metrics, the performance of the score function represented by AdaBoost sometimes surpasses that of the multi-head attention mechanism. This is possible because the multi-head attention mechanism contains more parameters and the current dataset used for training is relatively small, preventing the multi-head attention mechanism from demonstrating superior performance in exploring multiple feature dependencies. Furthermore, within the same method, the values of Hit@1, Hit@3, and Hit@5 increase sequentially. In the case of Hit@5, some methods, such as AdaBoost and the multi-head attention mechanism, exhibit extremely high hit rates, indicating a high probability of including the optimal model in SMAP’s recommendations when using these methods as score functions. However, this does not exclude the possibility that the top-1 model may have poor comprehensive performance. 
Commonly used recommendation algorithms such as SVD, SVD++, NMF, Slope One, and Co-Clustering have not demonstrated strong performance. This is because these methods rely solely on the scenarios, datasets, candidate models, and ratings for training, without taking into account the performance and characteristics of the model, thereby failing to comprehensively evaluate the model. However, among them, Slope One algorithm stands out for its high execution efficiency and ability to discover potential model preferences in the scenarios, making it perform better than the other recommendation algorithms mentioned above.

\begin{table*}[t]
    \begin{center}
    \caption{The optimal allocation between traffic scenarios and models.}\label{Allocation}
    \newcommand{\tabincell}[2]{\begin{tabular}{@{}#1@{}}#2\end{tabular}}
    \scalebox{0.8}{
    \begin{tabular}{c|c|c|c|c|c|c|c|c}
    \hline
    Traffic scenarios & Datasets& Models& Citations& Github stars& MAE& RMSE& MAPE&Optimal allocation \\ \hline
        \multirow{6}{*}{\tabincell{c}{Traffic speed \\ prediction}} &  \multirow{6}{*}{METR\_LA}& MTGNN & 324 & 487 & 2.76 & 5.34 & 5.18\% & \checkmark  \\ 
        && GGRU & 437 & 40 & 2.71 & 5.24 & 6.99\% & ~ \\ 
        && STGNN & 161 & 38 & 2.62 & 4.99 & 6.55\% & ~ \\ 
        && GTS & 37 & 113 & 3.01 & 5.85 & 8.20\% & ~ \\ 
        && ... &... &... & ... & ... & ... & ~ \\ 
        && HGCN & 22 & 64 & 2.89 & 5.51 & 7.48\%& ~ \\ \hline
        \multirow{6}{*}{\tabincell{c}{Road traffic \\ flow prediction}} &  \multirow{6}{*}{PEMSD4}& STFGNN & 119 & 124
 & 19.83 & 31.88 & 13.02\%&  ~  \\ 
        && T-GCN & 796 & 765 & 22.23 & 33.34 & 17.95\%& ~  \\ 
        && STGCN & 1668 & 662 & 18.85 & 30.00 & 13.09\%& \checkmark  \\ 
        && ASTGCN & 781 & 302 & 19.82 & 31.98 & 14.33\%& ~  \\ 
        && ... & ... & ... & ... & ... & ...& ~  \\  
        && TGC-LSTM & 521 & 294 & 22.03 & 34.23 & 16.41\%& ~  \\ \hline
        \multirow{6}{*}{\tabincell{c}{Station-level \\bus  passenger \\flow prediction}} &  \multirow{6}{*}{\tabincell{c}{Bus \\
        Transaction \\ Dataset}}& ST-GCN & 2335 & 107& 21.05 & 40.93 & 36.12\%& \checkmark  \\ 
        && GAT & 4447 & 2500 & 26.40 & 52.73 & 46.88\%& ~  \\ 
        && MS-Net & 3 & 0 & 19.15 & 36.42 & 33.12\%& ~ \\ 
        && DCRNN & 1588 & 864 & 30.37 & 51.37& 33.84\%& ~  \\ 
        && ... &... &... & ... & ... &  ~ \\ 
        && STGCN & 1668 & 662 & 32.52 & 54.33 & 38.41\%& ~  \\ \hline
        \multirow{6}{*}{\tabincell{c}{Ride-hailing \\demand prediction}} &  \multirow{6}{*}{TaxiNYC}& ARIMA &392 & 163& 16.01 & 34.68 & 22.55\%&  ~  \\ 
        && Multi-GCN & 187 & 17 & 10.77 & 26.12 & 32.42\%& ~  \\ 
        && DCRNN & 1588 & 864 & 11.23 & 27.11 & 30.88\%& \checkmark  \\ 
        && ConvLSTM & 5774 & 1200 & 12.69 & 29.56 & 35.57\%&  ~ \\ 
        && ... &... &... & ... & ... & ... &  ~ \\ 
        && AdaBoost & 122 & 0 & 14.27 & 31.74 & 19.28\%& ~  \\ \hline
        \multirow{6}{*}{\tabincell{c}{Station-level \\subway  passenger \\flow prediction}} &  \multirow{6}{*}{\tabincell{c}{Subway \\Transaction \\ Dataset}}& GAT & 4447 & 2500& 36.68 & 65.35 & 28.97\%& ~   \\ 
        && GSTNet & 89 & 9 & 21.33 & 36.08 & 18.63\%& ~  \\ 
        && ASTGCN & 781 & 302 & 31.40 & 47.85 & 19.57\%&  ~  \\ 
        && STGCN & 1668 & 662 & 31.16 & 46.81 & 19.65\%&  ~ \\ 
        && ... &... &... & ... & ... &  ~ \\ 
        && DCRNN & 1588 & 864 & 22.49 & 38.63 & 19.50\%& \checkmark  \\ \hline
        \multirow{6}{*}{\tabincell{c}{Taxi demand \\prediction}} &  \multirow{6}{*}{TaxiNYC}& LSTM & 72747 & 1500& 2.93 & 13.08 & 11.49\%& ~   \\ 
        && STGCN & 1668 & 662 & 2.01 & 4.03 & 7.02\%& ~  \\ 
        && GraphWaveNet & 607 & 365 & 2.02 & 4.13 & 7.19\%& ~  \\ 
        && STG2Seq & 118 & 24 & 1.94 & 4.04 & 6.74\%& ~ \\ 
        && ... &... &... & ... & ... & ... &~ \\ 
        && DCRNN & 1588 & 864 & 1.88 &3.66 & 6.42\%& \checkmark  \\ \hline
    \end{tabular}}
    \end{center}
\end{table*}

\subsubsection{Evaluation of the SOMA Algorithm}
The SOMA algorithm aims to achieve the optimal model allocation for new scenarios. In our SMAP, six popular traffic scenarios are selected as study cases, i.e., traffic speed prediction, road traffic flow prediction, station-level subway passenger flow prediction, station-level bus passenger flow prediction, taxi demand prediction and ride-hailing demand prediction. The allocation results shown in Table \ref{Allocation} demonstrate the effectiveness of the SOMA algorithm. The MAE, MAPE, and RMSE are three widely used metrics for evaluating traffic models. The selected suitable datasets and the candidate models that satisfy the constraints proposed in section definition $6$ are also listed in Table \ref{Allocation}.  For example, in road traffic flow prediction, SMAP selects PEMSD4 as the suitable dataset, and the candidate models include the STFGNN, T-GCN, STGCN, etc. Due to its high number of citations, numerous Github stars, and good performance, the STGCN as the optimal model is assigned to the road traffic flow prediction scenario by the SOMA algorithm. Its superior capacity to extract rich spatio-temporal dependencies from road networks helps DCRNN perform better and receive more attention from researchers. Therefore, DCRNN has the highest score computed by the score function and is assigned to the taxi demand prediction scenario as the optimal model. Although STGNN and GGRU have demonstrated excellent performance in traffic speed prediction, their limited number of GitHub stars may indicate that they could be challenging to reproduce. Therefore, they may not be the optimal models for this task. Models such as DCRNN, used for ride-hailing demand prediction and station-level subway passenger flow prediction, and ST-GCN, used for station-level bus passenger flow prediction, may not have the best performance but have received high citations and GitHub stars, leading to the highest overall ranking. Therefore, these models are considered optimal choices based on the score function.

\begin{figure}[t]
\centering
\frenchspacing
\includegraphics[width=3.7in]{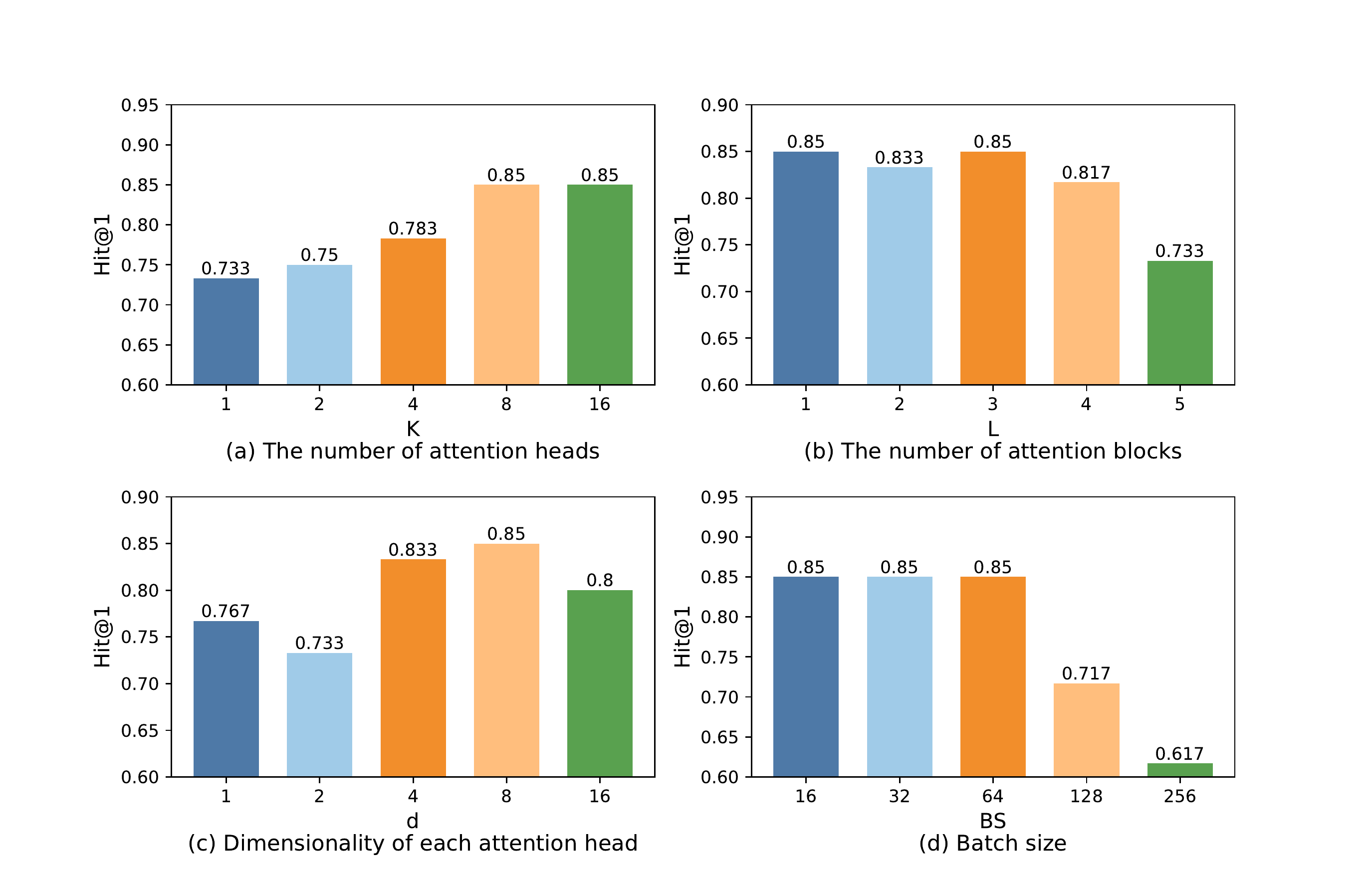}
\caption{Parameter study.}
\label{Parameter}
\end{figure}

\subsubsection{Parameter Study}
A parameter study is conducted on four hyper-parameters that can influence the effectiveness of the score function. These evaluated hyper-parameters include the number of attention heads $K$, the number of attention blocks $L$, the dimensionality $d$ of each attention head, and the batch size $BS$. Each experiment is repeated 10 times and report the average performance in Fig. \ref{Parameter}. The hyper-parameter under investigation is changed, and the other hyper-parameters are fixed in each experiment.
As Fig. \ref{Parameter} (a) shows, increasing the number of attention heads helps the model focus on different locations and improves its capacity to extract rich dependencies from all aspects, which initially enhances the model performance. However, the model faces overfitting when the number of attention heads is too large. Eight attention heads are sufficient for the score function to model and aggregate hidden correlations.
Fig. \ref{Parameter} (b) demonstrates that the performance of the score function is initially not sensitive to the number of attention blocks, but it became somewhat unstable under the parameter setting of 5. Moreover, the results indicate that a smaller number of attention blocks leads to higher Hit@1.
Fig. \ref{Parameter} (c) demonstrates that properly increasing the dimensionality of each attention head helps the model reserve the valuable information and dependencies extracted from different attention heads, which may enhance the model's expressive capacity. With the dimensionality of each attention head equals to 8, the score function has the highest Hit@1.
As depicted in Figure \ref{Parameter} (d), continuing to increase the batch size beyond a certain range leads to a more erroneous gradient descent direction, resulting in a greater training shock and weaker performance of the score function. In addition, selecting a batch size that is too large, such as $BS=128$ or $256$, can delay the parameter correction process and reduce the model’s generalization ability. However, it has been observed that a batch size of $64$ provides the optimal performance for the score function.

\section{Conclusion}
In this paper, a novel theoretical problem named the SOMA problem is developed, which aims to achieve the optimal model assignment by maximizing the total scores. To address this problem, an effective heterogeneous information framework entitled SMAP is proposed, which can integrate various types of information including scenarios, datasets and models, to automatically select a suitable dataset and efficiently assign the optimal model to a suitable dataset and a special scenario. Furthermore, SMAP develops a score function to comprehensively evaluate candidate models by adopting multi-head attention neural networks. The output of the score function includes the final score of each model. An SOMA algorithm based on the greedy approach is presented to allocate the optimal models to the newly added scenarios. A novel memory mechanism named the mnemonic center is developed to store the matched heterogeneous information and prevent duplicate matching; this mechanism can directly obtain the assigned scenario-dataset-model pairs without rematching. Full experiments conducted on a dataset demonstrate the effectiveness of the score function and SMAP. In the future, the score function and SOMA algorithm will be further optimized and the SMAP will be deployed in our cloud platform for accurate model assignment. 

\section*{Declaration of Interests}
The authors declare that they have no known competing financial interests or personal relationships that could have appeared to influence the work reported in this paper.

\section*{Authorship Credit and Contribution Statement}
Zekun Qiu: Conceptualization, Methodology, Writing - Original Draft, Experimental studies, Visualization. Zhipu Xie: Methodology, Writing - Original Draft. Zehua Ji: Supervision, Manuscript revision, Resources. Yuhao Mao: Supervision, Funding acquisition, Resources. Ke Cheng: Supervision, Manuscript revision, Resources.

\section*{Acknowledgement}
We would like to thank the anonymous reviewers for their constructive comments on this work. This work is supported by the National Natural Science Foundation of China through grants U1811463.

\appendix
\section{Typical Traffic Problems} \label{appendix1}

\begin{figure*}
	\centering
	\frenchspacing
\includegraphics[height=0.9\textheight,width=0.98\textwidth]{Traffic_Mind_Mapping.pdf}
\caption{Typical traffic problems.}
	\label{Traffic_mind_mapping}
\end{figure*} 

In this section, typical traffic problems are described, which can be divided into three types, i.e., traffic forecasting (green nodes), traffic anomaly detection (blue nodes) and route search and planning problems (yellow nodes). More details are shown in Fig. \ref{Traffic_mind_mapping}. 
Traffic forecasting problems are divided into different types for various transportation systems according to their predicted traffic states and list them, i.e., the prediction of traffic accidents, traffic speed, traffic congestion, occupancy, traffic flow, travel time, traffic demand and traffic density. Traffic speed, flow and demand predictions can be further subdivided based on specific scenarios. Two levels are contained in traffic speed prediction, namely, the road level and the regional level. In addition to the above two levels, traffic flow prediction also includes the station level.  The road-level flow prediction aims to predict traffic flows on roads and can be divided into road traffic flow, road OD flow, and intersection traffic throughput prediction. Different from the road-level flow prediction, the goal of the regional-level flow prediction is to predict the in- and out-volumes in each region of a city. According to the types of the provided services and predicted targets, the regional-level flow prediction can be partitioned into regional taxi flow, regional bike flow, regional riding-hailing flow, regional dockless e-scooter flow, regional OD taxi flow, regional OD bike flow, and regional OD ride-hailing flow prediction. The station-level flow prediction aims to predict the flow on a physical station, which includes station-level subway passenger flow, station-level bus passenger flow, station-level shared vehicle flow, station-level bike flow , and station-level railway passenger flow prediction.
Similar to the regional-level flow prediction, the traffic demand prediction can also be partitioned into taxi demand, bike demand, ride-hailing demand, and shared vehicle demand prediction. 
According to the needs of users, route search and planning problems can be divided into the following categories: route search, trajectory search, keyword-aware route planning, and location-based route planning. Route search can be further divided into location-based route search and keyword-aware route search based on the data types of user-defined entities. Trajectory search comprises trajectory similarity measure, trajectory-to-trajectory search, and trajectory-to-route search, each of which aims to finish the corresponding trajectory analysis task. Existing studies on keyword-aware route planning can be summarized into two types: exact matching-based and approximate matching-based researches. Exact matching-based keyword-aware route planning can also be divided into several subproblems, i.e., trip planning query, optimal sequenced route search, keyword-aware optimal route search, interactive route search, and group trip planning. The remaining issues are marked as other related issues. Route planning based on locations has attracted much research attention for a long time and can be divided into the following three subproblems: source-destination based route planning, multi-location based route planning, and multi-user based route planning. Six categories of related studies are summarized to represent the subproblems of source-destination based route planning: the time-dependent shortest problem, traffic-aware route planning, self-aware route planning, user preference based route planning, route planning over stochastic road network, and road planning over probabilistic road network. Multi-location based route planning includes three types of problems, i.e., mobile sequential recommendation, trajectory search by locations, and collective travel planning. Previous multi-user based routing planning studies can also be classified into two types: group based optimal route planning and global route planning.
Traffic anomaly detection is another challenging problem that has received increasing attention for decades. According to the types of processed data, traffic anomaly detection can be partitioned into time series anomaly detection, video-level anomaly detection, and image-level anomaly detection. Each category represents a special traffic scenario, and several related studies are listed in each category to demonstrate the accuracy of our classification process. Fig. \ref{Traffic_mind_mapping} also shows that a large number of traffic scenarios and numerous traffic models are presented. It is nontrivial to assign the optimal model to a special traffic scenario.

\section{Development of Machine Learning Relative to ITSs} \label{appendix2}
\begin{figure*}
	\centering
	\frenchspacing
\includegraphics[width=5.3in]{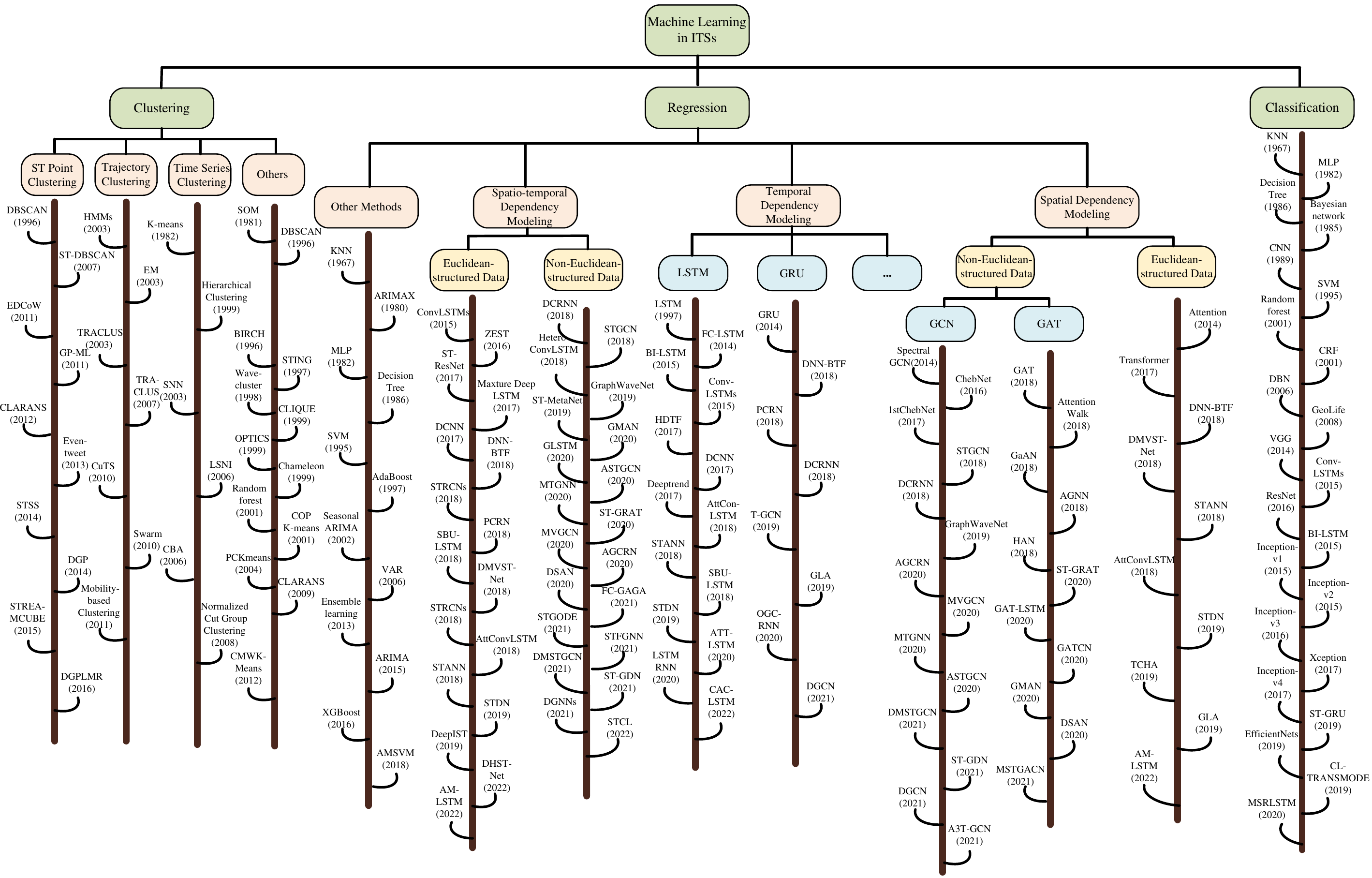}
\caption{Development of machine learning with respect to ITSs.}
	\label{MLTree}
\end{figure*}
ITSs are indispensable parts of smart cities that aim to improve the efficiency and security of transportation networks, which produce a great amount of transportation data every day. Machine learning has played an essential role in ITSs duo to its rapid development and strong capability to extract hidden correlations from big data.
Fig. \ref{MLTree} shows the development of machine learning with respect to ITSs. Clustering, regression, and classification are three types of methods that are widely used in various transportation tasks. According to the types of the processed data, clustering methods can be divided into ST point, trajectory, and time series clustering. The remaining issues are marked as other related issues. Regression methods are mainly utilized in traffic forecasting tasks and can be partitioned into spatial dependency modeling, temporal dependency modeling and spatio-temporal dependency modeling approaches. In spatial and spatio-temporal dependency modeling, two categories are summarized, i.e., the methods exploiting Euclidean-structured data and non-Euclidean-structured data. Graph convolutional neural networks (GCNs) and GATs are two major types of methods for handling non-Euclidean-structured data in spatial dependency modeling. A tree is constructed to present the development of each method, where each leaf describes the corresponding model name and the time of publication. Although some models were not initially designed for transportation, they have been widely used by many researchers to solve various traffic problems.
This is possibly because most of them are generic frameworks. For example, Inception-v1,v2,v3, and v4 \cite{Szegedy2015CVPR,pmlrv37ioffe15,Szegedy2016CVPR,10.5555/3298023.3298188}, which were proposed to improve the performance of deep neural networks and achieve accurate classification, are also applicable to some traffic tasks, such as image-level and video-level traffic anomaly detection. As can be seen, many traffic models are available. How to select the optimal model for a special traffic scenario is still a great challenge in the data mining community.

\bibliographystyle{elsarticle-num} 
\bibliography{DMAPBIB.bib}




\end{document}